\documentclass[11pt,a4paper]{article}
\usepackage[hyperref]{acl2020}

\usepackage{times}
\usepackage{latexsym}
\usepackage{graphicx}
\usepackage{lipsum}
\usepackage{boldline}
\usepackage{multirow}
\usepackage{lscape}
\usepackage{pbox}
\usepackage{tabulary}
\usepackage{amsmath}
\usepackage{array,amssymb}
\usepackage{makecell}
\usepackage{booktabs}
\usepackage{xspace}

\usepackage{bm}
\newcommand{\clickbaitadj}{click-baity\xspace}
\newcommand{\clickbaitnoun}{clickbait\xspace}
\newcommand{\clickbaitnouncap}{Clickbait\xspace}

\usepackage{microtype}

\aclfinalcopy 

\title{
Hooks in the Headline: Learning to Generate Headlines \\ with Controlled Styles
}

\author{Di Jin,\textsuperscript{\rm 1} Zhijing Jin,\textsuperscript{\rm 2} Joey Tianyi Zhou,\textsuperscript{\rm 3\thanks{Corresponding author.} {}} Lisa Orii,\textsuperscript{\rm 4} Peter Szolovits\textsuperscript{\rm 1}\\  
\textsuperscript{\rm 1}CSAIL, MIT,
\textsuperscript{\rm 2}Amazon Web Services,
\textsuperscript{\rm 3}A*STAR, Singapore, 
\textsuperscript{\rm 4}Wellesley College\\
\texttt{\{jindi15,psz\}@mit.edu,
zhijing.jin@connect.hku.hk}
\\
\texttt{
zhouty@ihpc.a-star.edu.sg,
lorii@wellesley.edu
} 
}

\date{}

\begin{document}
\maketitle
\begin{abstract}
  Current summarization systems only produce plain, factual headlines, but do not meet the practical needs of creating memorable titles to increase exposure. We propose a new task, Stylistic Headline Generation (SHG), to enrich the headlines with three style options (humor, romance and clickbait), in order to attract more readers. With \textit{no} style-specific article-headline pair (only a standard headline summarization dataset and mono-style corpora), our method TitleStylist generates style-specific headlines by combining the summarization and reconstruction tasks into a multitasking framework. We also introduced a novel parameter sharing scheme to further disentangle the style from the text. Through both automatic and human evaluation, we demonstrate that TitleStylist can generate relevant, fluent headlines with three target styles: humor, romance, and \clickbaitnoun. The attraction score of our model generated headlines surpasses that of the state-of-the-art summarization model by $9.68\%$, and even outperforms human-written references.\footnote{Our code is available at \url{https://github.com/jind11/TitleStylist}.}

\end{abstract}

\section{Introduction} \label{sec:intro}

\begin{figure}[!t]
    \centering
    \includegraphics[width= \columnwidth]{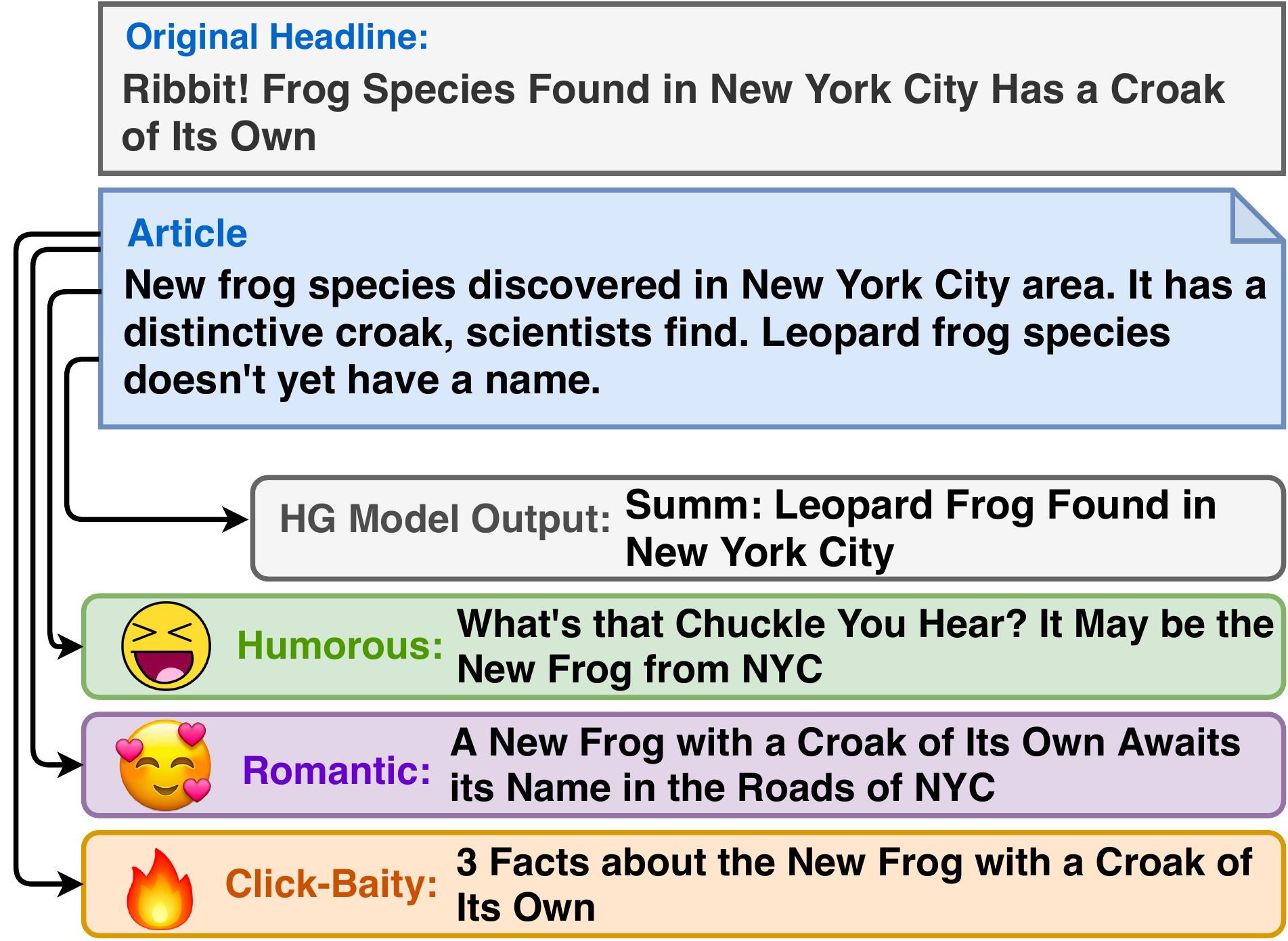}
    \caption{Given a news article, current HG models can only generate plain, factual headlines, failing to learn from the original human reference. It is also much less attractive than the headlines with humorous, romantic and \clickbaitadj styles.}
    \label{fig:intro-ex}
\end{figure}

Every good article needs a good title, which should not only be able to condense the core meaning of the text, but also sound appealing to the readers for more exposure and memorableness. However, currently even the best Headline Generation (HG) system can only fulfill the above requirement yet performs poorly on the latter. For example, in Figure~\ref{fig:intro-ex}, the plain headline by an HG model ``\textit{Summ:  Leopard Frog Found in New York City}'' is less eye-catching than the style-carrying ones such as ``\textit{What's That Chuckle You Hear? It May Be the New Frog From NYC}.''

To bridge the gap between the practical needs for attractive headlines and the plain HG by the current summarization systems, we propose a new task of Stylistic Headline Generation (SHG). Given an article, it aims to generate a headline with a target style such as humorous, romantic, and \clickbaitadj. It has broad applications in reader-adapted title generation, slogan suggestion, auto-fill for online post headlines, and many others.

SHG is a highly skilled creative process, and usually only possessed by expert writers. One of the most famous headlines in American publications, ``\textit{Sticks Nix Hick Pix},'' could be such an example. In contrast, the current best summarization systems are at most comparable to novice writers who provide a plain descriptive representation of the text body as the title \cite{cao2018faithful,Cao2018RetrieveRA,Lin2018GlobalEF,song2019mass,dong2019unified}. These systems usually use a language generation model that mixes styles with other linguistic patterns and inherently lacks a mechanism to control the style explicitly. More fundamentally, the training data comprise of a mixture of styles (e.g., the Gigaword dataset \cite{rush2017neural}), obstructing the models from learning a distinct style. 

In this paper, we propose the new task SHG, to emphasize the explicit control of style in headline generation. We present a novel headline generation model, TitleStylist, to produce enticing titles with target styles including humorous, romantic, and \clickbaitadj. Our model leverages a multitasking framework to train both a summarization model on headline-article pairs, and a Denoising Autoencoder (DAE) on a style corpus.
In particular, based on the transformer architecture \cite{vaswani2017attention}, we use the style-dependent layer normalization and the style-guided encoder-attention to disentangle the language style factors from the text. This design enables us to use the shared content to generate headlines that are more relevant to the articles, as well as to control the style by plugging in a set of style-specific parameters. We validate the model on three tasks: humorous, romantic, and \clickbaitadj headline generation. Both automatic and human evaluations show that TitleStylist can generate headlines with the desired styles that appeal more to human readers, as in Figure \ref{fig:intro-ex}.

The main contributions of our paper are listed below:
\begin{itemize}
    \item To the best of our knowledge, it is the first research on the generation of attractive news headlines with styles without any supervised style-specific article-headline paired data.
    \item Through both automatic and human evaluation, we demonstrated that our proposed  TitleStylist can generate relevant, fluent headlines with three styles (humor, romance, and \clickbaitnoun), and they are even more attractive than human-written ones.
    \item Our model can flexibly incorporate multiple styles, thus efficiently and automatically providing humans with various creative headline options for references and inspiring them to think out of the box.
\end{itemize}

\section{Related Work}

Our work is related to summarization and text style transfer.

\subsection*{Headline Generation as Summarization}
Headline generation is a very popular area of research. Traditional headline generation methods mostly focus on the extractive strategies using linguistic features and handcrafted rules~\citep{luhn1958automatic,edmundson1964problems,mathis1973improvement,salton1997automatic,jing1999decomposition,radev1998generating,dorr2003hedge}. To enrich the diversity of the extractive summarization, abstractive models were then proposed. With the help of neural networks, \citet{rush2015neural} proposed attention-based summarization (ABS) to make \citet{banko2000headline}'s framework of summarization more powerful. Many recent works extended ABS by utilizing additional features~\citep{chopra2016abstractive,takase2016neural,nallapati2016abstractive,shen2016neural,shen2017recent,tan2017neural,guo2017conceptual}. Other variants of the standard headline generation setting include headlines for community question answering~\citep{higurashi2018extractive}, multiple headline generation~\citep{iwamamultiple}, user-specific generation using user embeddings in recommendation systems~\cite{liu18review}, bilingual headline generation~\citep{DBLP:journals/taslp/ShenCYLS18} and question-style headline generation~\citep{DBLP:conf/cikm/ZhangGFLXCC18}.

Only a few works have recently started to focus on increasing the attractiveness of generated headlines~\citep{DBLP:conf/aclnmt/FanGA18,Xu2019ClickbaitSH}. \citet{DBLP:conf/aclnmt/FanGA18} 
focuses on controlling several features of the summary text such as text length, and the style of two different news outlets, CNN and DailyMail. These controls serve as a way to boost the model performance, and the CNN- and DailyMail-style control shows a negligible improvement. \citet{Xu2019ClickbaitSH} utilized reinforcement learning to encourage the headline generation system to generate more sensational headlines via using the readers' comment rate as the reward, which however cannot explicitly control or manipulate the styles of headlines.
\citet{Shu2018DeepHG} proposed a style transfer approach to transfer a non-clickbait headline into a clickbait one. This method requires paired news articles-headlines data for the target style; however, for many styles such as humor and romance, there are no available headlines. Our model does not have this limitation, thus enabling transferring to many more styles.  

\subsection*{Text Style Transfer}

Our work is also related to text style transfer, which aims to change the style attribute of the text while preserving its content. First proposed by \citet{DBLP:conf/nips/ShenLBJ17}, it has achieved great progress in recent years~\cite{Xu2018UnpairedST,Lample2019MultipleAttributeTR,Zhang2018StyleTA,fu2018style,Jin2019UnsupervisedTA,DBLP:conf/nips/YangHDXB18,jin2020unsupervised}. 
However, all these methods demand a text corpus for the target style; however, in our case, it is expensive and technically challenging to collect news headlines with humor and romance styles, which makes this category of methods not applicable to our problem.

\section{Methods}

\subsection{Problem Formulation}

The model is trained on a source dataset $S$ and target dataset $T$. The source dataset $S=\{(\bm{a^{(i)}},\bm{h^{(i)}})\}_{i=1}^N$ consists of pairs of a news article $\bm{a}$ and its \textit{plain} headline $\bm{h}$. We assume that the source corpus has a distribution $P(A, H)$, where $A=\{\bm{a^{(i)}}\}_{i=1}^N$, and $H=\{\bm{h^{(i)}}\}_{i=1}^N$. 
The target corpus $T=\{\bm{t^{(i)}}\}_{i=1}^{M}$ comprises of sentences $\bm{t}$ written in a specific style (e.g., humor). We assume that it conforms to the distribution $P(T)$. 

Note that the target corpus $T$ only contains style-carrying sentences, not necessarily headlines --- it can be just book text. Also no sentence $\bm{t}$ is paired with a news article. Overall, our task is to learn the conditional distribution $P(T|A)$ using only $S$ and $T$. This task is fully unsupervised because there is \textit{no} sample from the joint distribution $P(A, T)$.

\subsection{Seq2Seq Model Architecture}

\begin{figure}[!htpb]
    \centering
    \includegraphics[width= \columnwidth]{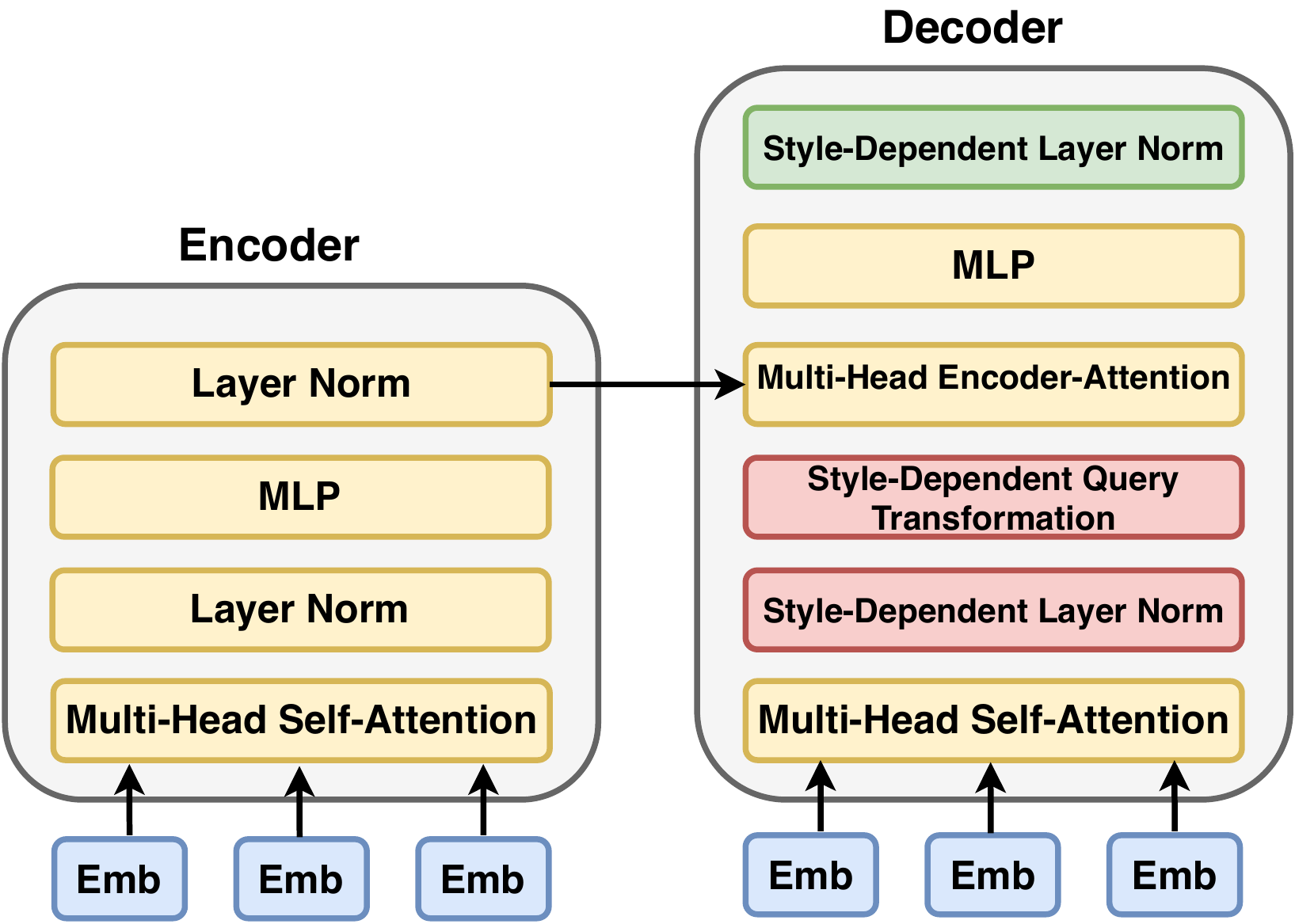}
    \caption{The Transformer-based architecture of our model.}
    \label{fig:model}
\end{figure}

For summarization, we adopt a sequence-to-sequence (Seq2Seq) model based on the Transformer architecture \cite{vaswani2017attention}. As in Figure \ref{fig:model}, it consists of a 6-layer encoder $E(\bm{\cdot}; \bm{\theta_E})$ and a 6-layer decoder $G(\bm{\cdot}; \bm{\theta_G})$ with a hidden size of 1024 and  a feed-forward filter size of 4096. For better generation quality, we initialize with the MASS model \citep{song2019mass}. MASS is pretrained by masking a sentence fragment in the encoder, and then predicting it in the decoder on large-scale English monolingual data. This pretraining is adopted in the current state-of-the-art systems across various summarization benchmark tasks including HG. 

\subsection{Multitask Training Scheme}

\begin{figure}[!htpb]
    \centering
    \includegraphics[width= \columnwidth]{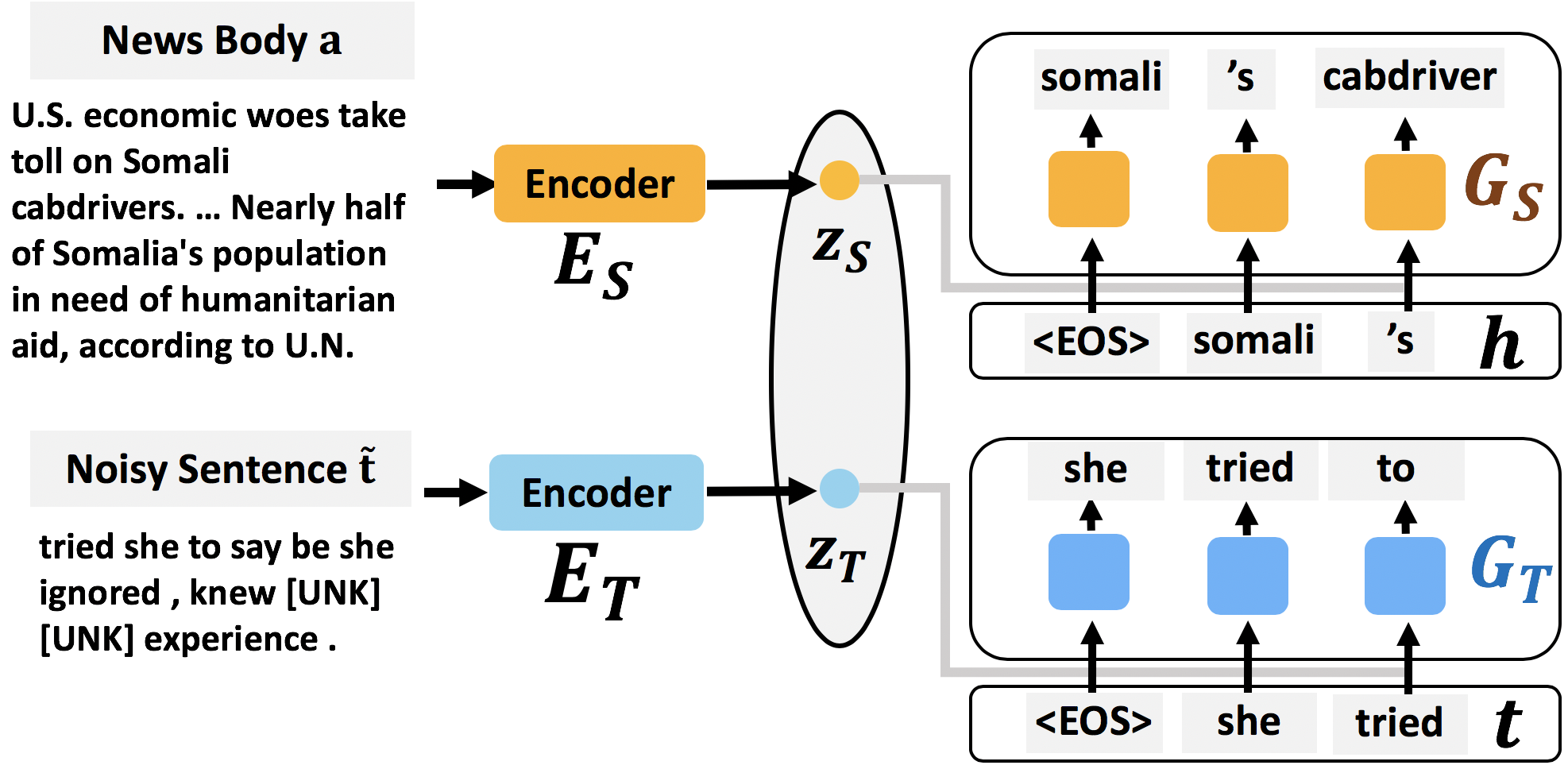}
    \caption{Training scheme. Multitask training is adopted to combine the summarization and DAE tasks.}
    \label{fig:training}
\end{figure}

To disentangle the latent style from the text, we adopt a multitask learning framework~\cite{Luong2015MultitaskST}, training on summarization and DAE simultaneously (as shown in Figure \ref{fig:training}).

\paragraph{Supervised Seq2Seq Training for $E_S$ and $G_S$}
With the source domain dataset $S$, based on the encoder-decoder architecture, we can learn the conditional distribution $P(H|A)$ by training $\bm{z}_S=E_S(A)$ and $H_S=G_S(\bm{z_S})$ to solve the supervised Seq2Seq learning task, where $\bm{z_S}$ is the learned latent representation in the source domain. The loss function of this task is
\begin{equation}
\small
    \mathcal{L}_S(\bm{\theta_{E_S}},\bm{\theta_{G_S}})=\mathbb{E}_{(\bm{a},\bm{h})\sim S} [-\log p(\bm{h}|\bm{a}; \bm{\theta_{E_S}},\bm{\theta_{G_S}})],
\end{equation}
where $\bm{\theta_{E_S}}$ and $\bm{\theta_{G_S}}$ are the set of model parameters of
the encoder and decoder in the source domain and $p(\bm{h}|\bm{a})$ denotes the overall probability of generating an output sequence $\bm{h}$ given the input article $\bm{a}$, which can be further expanded as follows:
\begin{equation}
\small
    p(\bm{h}|\bm{a}; \bm{\theta_{E_S}},\bm{\theta_{G_S}})=\prod_{t=1}^L p(h_t|\{h_1, ..., h_{t-1}\}, \bm{z_S}; \bm{\theta_{G_S}}),
\end{equation}
where $L$ is the sequence length.

\paragraph{DAE Training for $\bm{\theta_{E_T}}$ and $\bm{\theta_{G_T}}$}

For the target style corpus $T$, since we only have the sentence $\bm{t}$ without paired news articles, we train $\bm{z_T}=E_T(\bm{\tilde{t}})$ and $\bm{t}=G_T(\bm{z_T})$ by solving an unsupervised reconstruction learning task, where $\bm{z_T}$ is the learned latent representation in the target domain, and $\bm{\tilde{t}}$ is the corrupted version of $\bm{t}$ by randomly deleting or blanking some words and shuffling the word orders. To train the model, we minimize the reconstruction error $\mathcal{L}_T$:
\begin{equation}
    \mathcal{L}_T (\bm{\theta_{E_T}}, \bm{\theta_{G_T}})=\mathbb{E}_{\bm{t} \sim \bm{T}} [ - \log p(\bm{t}|\bm{\tilde{t}})],
\end{equation}
 where $\bm{\theta_{E_T}}$ and $\bm{\theta_{G_T}}$ are the set of model parameters for the encoder and generator in the target domain.
We train the whole model by jointly minimizing the supervised Seq2Seq training loss $\mathcal{L}_S$ and the unsupervised denoised auto-encoding loss $\mathcal{L}_T$ via multitask learning, so the total loss becomes

\begin{equation}
\begin{aligned}
    \mathcal{L}(\bm{\theta_{E_S}},\bm{\theta_{G_S}}, \bm{\theta_{E_T}},\bm{\theta_{G_T}})&=\lambda \mathcal{L}_S(\bm{\theta_{E_S}},\bm{\theta_{G_S}}) \\
    & + (1-\lambda)\mathcal{L}_T (\bm{\theta_{E_T}}, \bm{\theta_{G_T}}),
\end{aligned}
\end{equation}
where $\lambda$ is a hyper-parameter.

\subsection{Parameter-Sharing Scheme}

More constraints are necessary in the multitask training process. We aim to infer the conditional distribution as $
P(T|A)=G_T(E_S(A))$. However, without samples from $P(A, T)$, this is a challenging or even impossible task if $E_S$ and $E_T$, or $G_S$ and $G_T$ are completely independent of each other. Hence, we need to add some constraints to the network by relating $E_S$ and $E_T$, and $G_S$ and $G_T$. The simplest design is to share all parameters between $E_S$ and $E_T$, and apply the same strategy to $G_S$ and $G_T$. The intuition behind this design is that by exposing the model to both summarization task and style-carrying text reconstruction task, the model would acquire some sense of the target style while summarizing the article. However, to encourage the model to better disentangle the content and style of text and more explicitly learn the style contained in the target corpus $T$, we share all parameters of the encoder between two domains, i.e., between $E_S$ and $E_T$, whereas we divide the parameters of the decoder into two types: style-independent parameters $\bm{\theta_{\mathrm{ind}}}$ and style-dependent parameters $\bm{\theta_{\mathrm{dep}}}$. This means that only the style-independent parameters are shared between $G_S$ and $G_T$ while the style-dependent parameters are not. More specifically, the parameters of the layer normalization and encoder attention modules are made style-dependent as detailed below.

\paragraph{Type 1. Style Layer Normalization}
Inspired by previous work on image style transfer \cite{dumoulin2016learned}, we make the scaling and shifting parameters for layer normalization in the transformer architecture un-shared for each style. This \textit{style layer normalization} approach aims to transform a layer’s activation $\bm{x}$ into a normalized activation $\bm{z}$ specific to the style $s$:
\begin{equation}
    \bm{z}=\gamma_s(\frac{\bm{x}-\mu}{\sigma})-\beta_s,
\end{equation}
where $\mu$ and $\sigma$ are the mean and standard deviation of the batch of $\bm{x}$, and $\gamma_s$ and $\beta_s$ are style-specific parameters learned from data.

Specifically, for the transformer decoder architecture, we use a style-specific self-attention layer normalization and final layer normalization for the source and target domains on all six decoder layers. 

\paragraph{Type 2. Style-Guided Encoder Attention}
Our model architecture contains the attention mechanism, where the decoder infers the probability of the next word not only conditioned on the previous words but also on the encoded input hidden states. The attention patterns should be different for the summarization and the reconstruction tasks due to their different inherent nature. We insert this thinking into the model by introducing the \textit{style-guided encoder attention} into the multi-head attention module, which is defined as follows:

\begin{align}
& \bm{Q} = \bm{\mathrm{query}}\cdot \bm{W_q^s} \\
& \bm{K} = \bm{\mathrm{key}}\cdot \bm{W_k} \\
& \bm{V} = \bm{\mathrm{value}}\cdot \bm{W_v} \\
& \mathrm{Att}(\bm{Q},\bm{K},\bm{V})=\mathrm{Softmax}\left(\frac{\bm{Q}\bm{K^{\mathrm{tr}}}}{\sqrt{d_{\mathrm{model}}}}\right)\bm{V},
\end{align}

where $\bm{\mathrm{query}}$, $\bm{\mathrm{key}}$, and $\bm{\mathrm{value}}$ denote the triple of inputs into the multi-head attention module; $\bm{W_q^s}$, $\bm{W_k}$, and $\bm{W_v}$ denote the scaled dot-product matrix for affine transformation; $d_{\mathrm{model}}$ is the dimension of the hidden states. We specialize the dot-product matrix $\bm{W_q^s}$ of the query for different styles, so that $\bm{Q}$ can be different to induce diverse attention patterns. 

\section{Experiments}
\subsection{Datasets}
We compile a rich source dataset by combining the New York Times (NYT) and CNN, as well as three target style corpora on humorous, romantic, and \clickbaitadj text. The average sentence length in the NYT, CNN, Humor, Romance, and \clickbaitnouncap datasets are 8.8, 9.2, 12.6, 11.6 and 8.7 words, respectively.

\subsubsection{Source Dataset}
The source dataset contains news articles paired with corresponding headlines. To enrich the training corpus, we combine two datasets: the New York Times (56K) and CNN (90K). 
After combining these two datasets, we randomly selected 3,000
pairs as the validation set and another 3,000 pairs as the test set. 

We first extracted the archival abstracts and headlines from the New York Times (NYT) corpus~\cite{sandhaus2008new} and treat the abstracts as the news articles. Following the standard pre-processing procedures \cite{kedzie2018content},\footnote{\url{https://github.com/kedz/summarization-datasets}} we filtered out advertisement-related articles (as they are very different from news reports), resulting in 56,899 news abstracts-headlines pairs.

We then add into our source set the CNN summarization dataset, which is widely used for training abstractive summarization models~\cite{hermann2015teaching}.\footnote{We use CNN instead of the DailyMail dataset since DailyMail headlines are very long and more like short summaries.} We use the short summaries in the original dataset as the news abstracts and automatically parsed the headlines for each news from the dumped news web pages,\footnote{\url{https://cs.nyu.edu/~kcho/DMQA/}}  and in total collected 90,236 news abstract-headline pairs.

\subsubsection{Three Target Style Corpora}\label{sec:corpora}

\paragraph{Humor and Romance} For the target style datasets, we follow \cite{DBLP:conf/aaai/ChenP0S19} to use humor and romance novel collections in BookCorpus~\cite{zhu2015aligning} as the Humor and Romance datasets.\footnote{\url{https://www.smashwords.com/}} We 
split the documents into sentences, tokenized the text, and collected 500K sentences as our datasets.

\paragraph{Clickbait} We also tried to learn the writing style from the \clickbaitadj headlines since they have shown superior attraction to readers. Thus we used \textit{The Examiner - SpamClickBait News} dataset, denoted as the \clickbaitnouncap dataset.\footnote{\url{https://www.kaggle.com/therohk/examine-the-examiner}} We collected 500K headlines for our use.

Some examples from each style corpus are listed in Table~\ref{tab:datasets-examples}.

\begin{table}[htp]
\centering
\small
\begin{tabular}{p{1cm} p{6cm}}
\toprule
\textbf{Style}                      & \textbf{Examples} \\ \midrule
\multirow{6}{*}{Humor}      & - The crowded beach like houses in the burbs and the line ups at Walmart.                 \\
                           & - Berthold stormed out of the brewing argument with his violin and bow and went for a walk with it to practice for the much more receptive polluted air.         \\ \midrule
\multirow{6}{*}{Romance}     & - ``I can face it joyously and with all my heart, and soul!'' she said.      \\ 
                           & - With bright blue and green buttercream scales, sparkling eyes, and purple candy melt wings, it sat majestically on a rocky ledge made from chocolate.  \\\midrule
\multirow{6}{*}{\clickbaitnouncap}  & - 11-Year-Old Girl and 15-Year-Old Boy Accused of Attempting to Kill Mother: Who Is the Adult? \\
                        & - Chilly, Dry Weather Welcomes 2010 to South Florida       \\
                           & - End Segregation in Alabama-Bryce Hospital Sale Offers a Golden Opportunity  \\ \bottomrule 
\end{tabular}
\caption{Examples of three target style corpora: humor, romance, and \clickbaitnoun.}
\label{tab:datasets-examples}
\end{table}


\subsection{Baselines}

We compared the proposed TitleStylist against the following five strong baseline approaches.

\paragraph{Neural Headline Generation (NHG)}
We train the state-of-the-art summarization model, MASS~\cite{song2019mass}, on our collected news abstracts-headlines paired data.

\paragraph{Gigaword-MASS}
We test an off-the-shelf headline generation model, MASS from \cite{song2019mass}, which is already trained on Gigaword, a large-scale headline generation dataset with around 4 million articles.\footnote{\url{https://github.com/harvardnlp/sent-summary}}

\paragraph{Neural Story Teller (NST)}
It breaks down the task into two steps, which first generates headlines from the aforementioned NHG model, then applies style shift techniques to generate style-specific headlines~\cite{kiros2015skip}. In brief, this method uses the Skip-Thought model to encode a sentence into a representation vector and then manipulates its style by a linear transformation. Afterward, this transformed representation vector is used to initialize a language model pretrained on a style-specific corpus so that a stylistic headline can be generated. More details of this method can refer to the official website.\footnote{\url{https://github.com/ryankiros/neural-storyteller}}

\paragraph{Fine-Tuned}
We first train the NHG model as mentioned above, then further fine-tuned it on the target style corpus via DAE training.

\paragraph{Multitask}
We share all parameters between $E_S$ and $E_T$, and between $G_S$ and $G_T$, and trained the model on both the summarization and DAE tasks. The model architecture is the same as NHG.

\subsection{Evaluation Metrics}
To evaluate the performance of the proposed TitleStylist in generating attractive headlines with styles, we propose a comprehensive twofold strategy of both automatic evaluation and human evaluation.

\subsubsection{Setup of Human Evaluation}

We randomly sampled 50 news abstracts from the test set and asked three native-speaker annotators for evaluation to score the generated headlines. Specifically, we conduct two tasks to evaluate on four criteria: (1) relevance, (2) attractiveness, (3) language fluency, and (4) style strength. For the first task, the human raters are asked to evaluate these outputs on the first three aspects, relevance, attractiveness, and language fluency on a Likert scale from $1$ to $10$ (integer values). For \textbf{relevance}, human annotators are asked to evaluate how semantically relevant the headline is to the news body. For \textbf{attractiveness}, annotators are asked how attractive the headlines are. For \textbf{fluency}, we ask the annotators to evaluate how fluent and readable the text is. After the collection of human evaluation results, we averaged the scores as the final score. In addition, we have another independent human evaluation task about the \textbf{style strength} -- we present the generated headlines from TitleStylist and baselines to the human judges and let them choose the one that most conforms to the target style such as humor. Then we define the style strength score as the proportion of choices. 

\subsubsection{Setup of Automatic Evaluation}

Apart from the comprehensive human evaluation, we use automatic evaluation to measure the generation quality through  two conventional aspects: summarization quality and language fluency. Note that the purpose of this two-way automatic evaluation is to confirm that the performance of our model is in an acceptable range. Good automatic evaluation performances are necessary proofs to compliment human evaluations on the model effectiveness.

\paragraph{Summarization Quality}
We use the standard automatic evaluation metrics for summarization with the original headlines as the reference: BLEU~\cite{papineni2002bleu}, METEOR~\cite{denkowski2014meteor}, ROUGE~\cite{lin2004rouge} and CIDEr~\cite{vedantam2015cider}. For ROUGE, we used the Files2ROUGE\footnote{\url{https://github.com/pltrdy/files2rouge}} toolkit, and for other metrics, we used the pycocoeval toolkit.\footnote{\url{https://github.com/Maluuba/nlg-eval}}

\paragraph{Language Fluency}
We fine-tuned the GPT-2 medium model~\cite{Radford2019LanguageMA} on our collected headlines and then used it to measure the perplexity (PPL) on the generated outputs.\footnote{PPL on the development set is 42.5}

\subsection{Experimental Details}

We used the fairseq code base~\cite{ott2019fairseq}.
During training, we use Adam optimizer with an initial learning rate of $5\times10^{-4}$, and the batch size is set as 3072 tokens for each GPU with the parameters update frequency set as 4. 
For the random corruption for DAE training, we follow the standard practice to randomly delete or blank the word with a uniform probability of $0.2$, and randomly shuffled the word order within $5$ tokens. All datasets are lower-cased. $\lambda$ is set as 0.5 in experiments. For each iteration of training, we randomly draw a batch of data either from the source dataset or from the target style corpus, and the sampling strategy follows the uniform distribution with the probability being equal to $\lambda$.

\section{Results and Discussion}

\subsection{Human Evaluation Results}

The human evaluation is to have a comprehensive measurement of the performances. We conduct experiments on four criteria, relevance, attraction, fluency, and style strength. We summarize the human evaluation results on the first three criteria in Table~\ref{tab:human-eval}, and the last criteria in Table~\ref{tab:style-strength}. Note that through automatic evaluation, the baselines NST, Fine-tuned, and Gigaword-MASS perform poorer than other methods (in Section ~\ref{sec:auto_eval}), thereby we removed them in human evaluation to save unnecessary work for human raters.

\begin{table}[!hptb]
\centering
\small
\resizebox{0.5\textwidth}{!}{\begin{tabular}{llccc}
\toprule
\textbf{Style}                      & \textbf{Settings}         & \textbf{Relevance} & \textbf{Attraction} & \textbf{Fluency} \\ \midrule
\multirow{2}{*}{None}      & NHG              & \textbf{6.21}      & 8.47       & 9.31    \\
                           & Human         & 5.89      & 8.93       & 9.33    \\ \midrule
\multirow{2}{*}{Humor}     & Multitask       & 5.51      & 8.61       & 9.11    \\
                           & TitleStylist & 5.87      & 8.93       & 9.29    \\\midrule
\multirow{2}{*}{Romance}   & Multitask       & 5.67      & 8.54       & 8.91    \\
                           & TitleStylist & 5.86      & 8.87       & 9.14    \\ \midrule
\multirow{2}{*}{Clickbait} & Multitask       & 5.67      & 8.71       & 9.21    \\
                           & TitleStylist & 5.83      & \textbf{9.29}       & \textbf{9.44}  \\ \bottomrule 
\end{tabular}}
\caption{Human evaluation on three aspects: relevance, attraction, and fluency. ``None'' represents the original headlines in the dataset.}
\label{tab:human-eval}
\end{table}

\begin{table*}[!t]
\centering
\small
\renewcommand{\arraystretch}{1.2}
\begin{tabular}{ >{\raggedright\arraybackslash} p{1.2cm} >{\arraybackslash} m{6.8cm}>{\arraybackslash} m{6.8cm}}
\toprule
\textbf{News Abstract}         & Turkey's bitter history with Kurds is figuring prominently in its calculations over how to deal with Bush administration's request to use Turkey as the base for thousands of combat troops if there is a war with Iraq; Recep Tayyip Erdogan, leader of Turkey's governing party, says publicly for the first time that future of Iraq's Kurdish area, which abuts border region of Turkey also heavily populated by Kurds, is weighing heavily on negotiations; Hints at what Turkish officials have been saying privately for weeks: if war comes to Iraq, overriding Turkish objective would be less helping Americans topple Saddam Hussein, but rather preventing Kurds in Iraq from forming their own state. & Reunified Berlin is commemorating 40th anniversary of the start of construction of Berlin wall, almost 12 years since Germans jubilantly celebrated reopening between east and west and attacked hated structure with sledgehammers; Some Germans are championing the preservation of wall at the time when little remains beyond few crumbling remnants to remind Berliners of unhappy division that many have since worked hard to heal and put behind them; What little remains of physical wall embodies era that Germans have yet to resolve for themselves; They routinely talk of 'wall in the mind' to describe social and cultural differences that continue to divide easterners and westerners. \\ \midrule
\textbf{Human} & Turkey assesses question of Kurds  & The wall Berlin can't quite demolish \\ 
\textbf{NHG}     & Turkey's bitter history with Kurds   &  Construction of Berlin wall is commemorated \\ \midrule
\textbf{Humor} & What if there is a war with Kurds?  & The Berlin wall, 12 years later, is still there? \\ 
\textbf{Romance} & What if the Kurds say ``No'' to Iraq? & The Berlin wall: from the past to the present \\ 
\textbf{Clickbait} & For Turkey, a long, hard road & East vs West, Berlin wall lives on \\ 
\bottomrule
\end{tabular}
\caption{Examples of style-carrying headlines generated by TitleStylist.}
\label{tab:model-examples}
\end{table*}

\paragraph{Relevance}
We first look at the relevance scores in Table~\ref{tab:human-eval}. It is interesting but not surprising that the pure summarization model NHG achieves the highest relevance score. The outputs from NHG are usually like an organic reorganization of several keywords in the source context (as shown in Table~\ref{tab:model-examples}), thus appearing most relevant. It is noteworthy that the generated headlines of our TitleStylist for all three styles are close to the original human-written headlines in terms of relevance, validating that our generation results are qualified in this aspect. Another finding is that more attractive or more stylistic headlines would lose some relevance since they need to use more words outside the news body for improved creativity.

\paragraph{Attraction}
In terms of attraction scores in Table~\ref{tab:human-eval}, we have three findings: (1) The human-written headlines are more attractive than those from NHG, which agrees with our observation in Section~\ref{sec:intro}. (2) Our TitleStylist can generate more attractive headlines over the NHG and Multitask baselines for all three styles, demonstrating that adapting the model to these styles could improve the attraction and specialization of some parameters in the model for different styles can further enhance the attraction. (3) Adapting the model to the ``\clickbaitnouncap'' style could create the most attractive headlines, even out-weighting the original ones, which agrees with the fact that \clickbaitadj headlines are better at drawing readers' attention. To be noted, although we learned the ``\clickbaitnouncap'' style into our summarization system, we still made sure that we are generating relevant headlines instead of too exaggerated ones, which can be verified by our relevance scores.

\paragraph{Fluency}
The human-annotated fluency scores in Table~\ref{tab:human-eval} verified that our TitleStylist generated headlines are comparable or superior to the human-written headlines in terms of readability. 

\paragraph{Style Strength}
We also validated that our TitleStylist can carry more styles compared with the Multitask and NHG baselines by summarizing the percentage of choices by humans for the most humorous or romantic headlines in Table~\ref{tab:style-strength}. 

\begin{table}[h]
\centering
\small
\resizebox{0.48\textwidth}{!}{\begin{tabular}{lccc}
\toprule
\textbf{Style}   & \textbf{NHG} & \textbf{Multitask} & \textbf{TitleStylist} \\ \midrule
Humor   & 18.7     & 35.3            & \textbf{46.0}               \\ 
Romance & 24.7     & 34.7            & \textbf{40.6}                \\ 
Clickbait & 13.8   & 35.8       & \textbf{50.4}                \\ \bottomrule
\end{tabular}}
\caption{Percentage of choices (\%) for the most humorous or romantic headlines among TitleStylist and two baselines NHG and Multitask.}
\label{tab:style-strength}
\end{table}

\subsection{Automatic Evaluation Results} \label{sec:auto_eval}

Apart from the human evaluation of the overall generation quality on four criteria, we also conducted a conventional automatic assessment to gauge only the summarization quality. This evaluation  does not take other measures such as the style strength into consideration, but it serves as important complimentary proof to ensure that the model has an acceptable level of summarization ability.

Table~\ref{tab:auto-eval} summarizes the automatic evaluation results of our proposed TitleStylist model and all baselines. We use the summarization-related evaluation metrics, i.e., BLEU, ROUGE, CIDEr, and METEOR, to measure how relevant the generated headlines are to the news articles, to some extent, by comparing them to the original human-written headlines. In Table~\ref{tab:auto-eval}, the first row ``NHG'' shows the performance of the current state-of-the-art summarization model on our data, and Table~\ref{tab:model-examples} provides two examples of its generation output. Our ultimate goal is to generate more attractive headlines than these while maintaining relevance to the news body.

\begin{table*}[t]
\centering
\small
\resizebox{1.\textwidth}{!}{\begin{tabular}{llcccccccc}
\toprule
\textbf{Style Corpus}                      & \textbf{Model}         & \textbf{BLEU} & \textbf{ROUGE-1} & \textbf{ROUGE-2} & \textbf{ROUGE-L} & \textbf{CIDEr} & \textbf{METEOR} & \textbf{PPL ($\downarrow$)}    & \textbf{Len. Ratio (\%)} \\ \midrule
\multirow{2}{*}{None}      & NHG              & 12.9 & 27.7    & 9.7     & 24.8    & 0.821 & 0.123  & 40.4  & 8.9        \\ 
                           & Gigaword-MASS    & 9.2  & 22.6    & 6.4     & 20.1    & 0.576 & 0.102  & 65.0  & 9.7        \\ \midrule
\multirow{5}{*}{Humor}     & NST              & 5.8  & 17.8    & 4.3     & 16.1    & 0.412 & 0.078  & 361.3 & 9.2        \\ 
                           & Fine-tuned       & 4.3  & 15.7    & 3.4     & 13.2    & 0.140 & 0.093  & 398.8 & 3.9        \\ 
                           & Multitask       & 14.7 & 28.9    & \textbf{11.6}    & 26.1    & 0.995 & 0.134  & 40.0  & 9.5        \\ 
                           & TitleStylist & 13.3 & 28.1    & 10.3    & 25.4    & 0.918 & 0.127  & 46.2  & 10.6       \\ 
                           & TitleStylist-F & \textbf{15.2} & \textbf{29.2}    & \textbf{11.6}    & \textbf{26.3}    & \textbf{1.022} & \textbf{0.135}  & \textbf{39.3}  & 9.7        \\ 
                           \midrule
\multirow{5}{*}{Romance}   & NST              & 2.9  & 9.8     & 0.9     & 9.0     & 0.110 & 0.047  & 434.1 & 6.2        \\ 
                           & Fine-tuned       & 5.1  & 18.7    & 4.5     & 16.1    & 0.023 & 0.128  & 132.2 & 2.8        \\ 
                           & Multitask       & 14.8 & 28.7    & 11.5    & 25.9    & 0.997 & 0.132  & 40.5  & 9.7        \\ 
                           
                           & TitleStylist & 12.0 & 27.2    & 10.1    & 24.4    & 0.832 & 0.134  & 40.1  & 7.4        \\
                           & TitleStylist-F & \textbf{15.0} & \textbf{29.0}    & \textbf{11.7}    & \textbf{26.2}    & \textbf{1.005} & \textbf{0.134}  & \textbf{39.0}  & 9.8        \\ 
                           \midrule
\multirow{5}{*}{Clickbait} & NST              & 2.5  & 8.4     & 0.6     & 7.8     & 0.089 & 0.041  & 455.4 & 6.3        \\ 
                           & Fine-tuned       & 4.7  & 17.3    & 4.0     & 15.0    & 0.019 & 0.116  & 172.0 & 2.8        \\ 
                           & Multitask       & 14.5 & 28.3    & 11.2    & 25.5    & 0.980 & 0.132  & \textbf{38.5}  & 9.7        \\
                           & TitleStylist & 11.5 & 26.6    & 9.8     & 23.7    & 0.799 & \textbf{0.134}  & 40.7  & 7.3        \\  
                           & TitleStylist-F & \textbf{14.7} & \textbf{28.6}    & \textbf{11.4}    & \textbf{25.9}    & \textbf{0.981} & 0.133  & 38.9  & 9.6        \\ \bottomrule
\end{tabular}}
\caption{Automatic evaluation results of our TitleStylist and baselines. The test set of each style is the same, but the training set is different depending on the target style as shown in the 
``Style Corpus'' column. ``None'' means no style-specific dataset, and ``Humor'', ``Romance'' and ``Clickbait'' corresponds to the datasets we introduced in Section~\ref{sec:corpora}.
During the inference phase, our TitleStylist can generate two outputs: one from $G_T$ and the other from $G_S$. Outputs from $G_T$ are style-carrying, so we denote it as ``TitleStylist''; outputs from $G_S$ are plain and factual, thus denoted as ``TitleStylist-F.'' The last column ``Len.~Ratio'' denotes the average ratio of abstract length to the generated headline length by the number of words.}
\label{tab:auto-eval}
\end{table*}

 From Table~\ref{tab:auto-eval}, the baseline Gigaword-MASS scored worse  than NHG, revealing that directly applying an off-the-shelf headline generation model to new in-domain data is not feasible, although this model has been trained on more than 20 times larger dataset. Both NST and Fine-tuned baselines present very poor summarization performance, and the reason could be that both of them cast the problem into two steps: summarization and style transfer, and the latter step is absent of the summarization task, which prevents the model from maintaining its summarization capability. 

In contrast, the Multitask baseline involves the summarization and style transfer (via reconstruction training) processes at the same time and shows superior summarization performance even compared with NHG. This reveals that the unsupervised reconstruction task can indeed help improve the supervised summarization task. More importantly, we use two different types of corpora for the reconstruction task: one consists of headlines that are similar to the news data for the summarization task, and the other consists of text from novels that are entirely different from the news data. However, unsupervised reconstruction training on both types of data can contribute to the summarization task, which throws light on the potential future work in summarization by incorporating unsupervised learning as augmentation.

We find that in Table~\ref{tab:auto-eval} TitleStylist-F achieves the best summarization performance. This implicates that, compared with the Multitask baseline where the two tasks share all parameters, specialization of layer normalization and encoder-attention parameters can make $G_S$ focus more on summarization. 

It is noteworthy that the summarization scores for TitleStylist are lower than TitleStylist-F but still comparable to NHG. This agrees with the fact that the $G_T$ branch more focuses on bringing in stylistic linguistic patterns into the generated summaries, thus the outputs would deviate from the pure summarization to some degree. However, the relevance degree of them remains close to the baseline NHG, which is the starting point we want to improve on. Later in the next section, we will further validate that these headlines are faithful to the new article through human evaluation.

We also reported the perplexity (PPL) of the generated headlines to evaluate the language fluency, as shown in Table~\ref{tab:auto-eval}. All outputs from baselines NHG and Multitask and our proposed TitleStylist show similar PPL compared with the test set (used in the fine-tuning stage) PPL 42.5, indicating that they are all fluent expressions for news headlines.

\subsection{Extension to Multi-Style}

We progressively expand TitleStylist to include all three target styles (humor, romance, and \clickbaitnoun) to demonstrate the flexibility of our model. That is, we simultaneously trained the summarization task on the headlines data and the DAE task on the three target style corpora. And we made the layer normalization and encoder-attention parameters specialized for these four styles (fact, humor, romance, and \clickbaitnoun) and shared the other parameters. We compared this multi-style version, TitleStylist-Versatile, with the previously presented single-style counterpart, as shown in Table~\ref{tab:multi-style}. From this table, we see that the BLEU and ROUGE-L scores of TitleStylist-Versatile are comparable to TitleStylist for all three styles. Besides, we conducted another human study to determine the better headline between the two models in terms of attraction, and we allow human annotators to choose both options if they deem them as equivalent. The result is presented in the last column of Table~\ref{tab:multi-style}, which shows that the attraction of TitleStylist-Versatile outputs is competitive to TitleStylist. TitleStylist-Versatile thus generates multiple headlines in different styles altogether, which is a novel and efficient feature.

\begin{table}[]
\centering
\small
\resizebox{0.49\textwidth}{!}{\begin{tabular}{llccc}
\toprule
\textbf{Style}                      & \textbf{Model}                & \textbf{BLEU} & \textbf{RG-L} & \textbf{Pref. (\%)} \\ \midrule
{None}                       & TitleStylist-Versatile &  14.5    &  25.8       &   ---    \\ \midrule
\multirow{2}{*}{Humor}     & TitleStylist-Versatile &  12.3    &      24.5   &   42.6    \\
                           & TitleStylist       &  \textbf{13.3}    &  \textbf{25.4}       &  \textbf{57.4}     \\ \midrule
\multirow{2}{*}{Romance}   & TitleStylist-Versatile &  \textbf{12.0}    &  24.2       &   46.3    \\
                           & TitleStylist       &  \textbf{12.0}    &      \textbf{24.4}   &  \textbf{53.7}     \\ \midrule
\multirow{2}{*}{Clickbait} & TitleStylist-Versatile &  \textbf{13.1}    &  \textbf{24.9}       &   \textbf{52.9}    \\
                           & TitleStylist       &  11.5    &  23.7       &  47.1
                           \\ \bottomrule
\end{tabular}}
\caption{Comparison between TitleStylist-Versatile and TitleStylist. ``RG-L'' denotes ROUGE-L, and ``Pref.'' denotes preference.}
\label{tab:multi-style}
\end{table}


\section{Conclusion}
We have proposed a new task of Stylistic Headline Generation (SHG) to emphasize explicit control of styles in headline generation for improved attraction. To this end, we presented a multitask framework to induce styles into summarization, and proposed the parameters sharing scheme to enhance both summarization and stylization capabilities. Through experiments, we validated our proposed TitleStylist can generate more attractive headlines than state-of-the-art HG models.

\section*{Acknowledgement}
We appreciate all the volunteer native speakers (Shreya Karpoor, Lisa Orii, Abhishek Mohan, Paloma Quiroga, etc.) for the human evaluation of our study, and thank the reviewers for their inspiring comments. Joey Tianyi Zhou is partially supported  by the Agency for Science, Technology and Research (A*STAR) under its AME Programmatic Funding Scheme (Project No. A18A1b0045).
\bibliography{acl2020}
\bibliographystyle{acl_natbib}

\end{document}